%% file: main.tex
\documentclass[11pt,a4paper]{article}
\usepackage[hyperref]{template_emnlp20/emnlp2020}
\usepackage{times}
\usepackage{latexsym}

\usepackage{hyperref}
\usepackage{url}
\usepackage{graphicx}
\usepackage{multirow}
\usepackage{arydshln}

\newcommand\blfootnote[1]{%
  \begingroup
  \renewcommand\thefootnote{}\footnote{#1}%
  \addtocounter{footnote}{-1}%
  \endgroup
}

\usepackage{microtype}

\aclfinalcopy 

\input{math_commands.tex}

\title{Structured Pruning of Large Language Models}

\author{Ziheng Wang \textsuperscript{*} \\
  ASAPP, Inc. \\
  \small{\texttt{zihengw@stanford.edu}} \\\And
  Jeremy Wohlwend \textsuperscript{*} \\
  ASAPP, Inc. \\
  \small{\texttt{jwohlwend@csail.mit.edu}} \\\And
  Tao Lei \textsuperscript{*} \\
  ASAPP, Inc. \\
  \small{\texttt{tao@asapp.com}} \\}

\begin{document}
\maketitle

\begin{abstract}
\input{text/abstract.tex}
\end{abstract}

\section{Introduction}
\input{text/introduction.tex}

\section{Related Work}
\input{text/related.tex}

\input{text/method.tex}

\input{text/results.tex}

\section{Analysis}
\input{text/analysis.tex}

\section{Conclusion}
\input{text/conclusion.tex}

\section*{Acknowledgement}
\input{text/acknowledgement.tex}

\bibliography{main}
\bibliographystyle{template_emnlp20/acl_natbib}

\appendix
\input{text/appendix.tex}

\end{document}

%% file: math_commands.tex
\usepackage{amsmath,amsfonts,bm}

\def\eqref#1{equation~\ref{#1}}

\def\1{\bm{1}}

\def\rmE{{\mathbf{E}}}

\def\rmG{{\mathbf{G}}}

\def\rmO{{\mathbf{O}}}

\def\rmW{{\mathbf{W}}}

\DeclareMathAlphabet{\mathsfit}{\encodingdefault}{\sfdefault}{m}{sl}
\SetMathAlphabet{\mathsfit}{bold}{\encodingdefault}{\sfdefault}{bx}{n}

\newcommand{\R}{\mathbb{R}}

\newcommand{\lzero}{$l_0$\ }
\newcommand{\lone}{$l_1$\ }
\newcommand{\mask}{\mathbf{z}}
\newcommand{\param}{\bm{\theta}}
\newcommand{\ourmethod}{FLOP }

%% file: text/abstract.tex
Large language models have recently achieved state of the art performance across a wide variety of natural language tasks. Meanwhile, the size of these models and their latency have significantly increased, which makes their usage costly, and raises an interesting question: do language models need to be large? We study this question through the lens of model compression. We present a generic, structured pruning approach by parameterizing each weight matrix using its low-rank factorization, and adaptively removing rank-1 components during training. On language modeling tasks, our structured approach outperforms other unstructured and block-structured pruning baselines at various compression levels, while achieving significant speedups during both training and inference. We also demonstrate that our method can be applied to pruning adaptive word embeddings in large language models, and to pruning the BERT model on several downstream fine-tuning classification benchmarks.\textsuperscript{1}
\blfootnote{\textsuperscript{*}Denotes equal contribution.}
\blfootnote{\textsuperscript{1}Our code is publicly available at \url{https://github.com/asappresearch/flop}.}

%% file: text/introduction.tex
Recent advances in language modeling have led to remarkable improvements on a variety of natural language tasks~\citep{dai2015semi,peters2018deep,radford2018improving,devlin2018bert,liu2019roberta,dai2019transformer,zhang2019ernie}. These models, however, have grown increasingly large, rendering them slow and expensive for real-world applications. Through the use of model compression, we aim to reduce this overhead, and to better understand the role of model capacity in large language models.

A common approach to model compression is known as weight pruning \citep{zhu2017prune,han2015deep,see2016compression}. Model weights are progressively removed, resulting in sparse matrices across the network. Earlier work focuses mostly on unstructured pruning, where weights are pruned individually \citep{narang2017exploring,zhu2017prune}. While this method is effective, it results in unstructured sparse matrices that are difficult to support on common hardware \citep{han2016eie}, making it challenging to obtain training and inference speedups despite a significant reduction in model size.

On the other hand, \emph{structured pruning} imposes structured sparse patterns by removing groups of consecutive parameters, such as rows, columns or $k\times k$ sub-blocks of the weight matrix~\citep{narang2017block,wen2017learning,cao2019efficient}.
These methods lead to significant speedup, but tend to give lower performance than unstructured pruning given the same parameter budget~\citep{yao2019balanced}.

Another caveat is that some of these methods require special linear algebra implementations~\citep{gray2017gpu,yao2019balanced} or hardware~\citep{cao2019efficient} in order to accelerate matrix multiplication, therefore limiting their application to a broad set of existing models.

We propose a generic, improved structured pruning approach based on adaptive low-rank factorization. As an alternative to unstructured sparse and block sparse representations, low-rank factorization retains the full dense structure of weight matrices, eliminating the need for special linear algebra primitives and hardware for computation speedup. Compared to row (and column) based pruning, low-rank factorization better preserves the linear transformation of the un-compressed matrices. During training, our method adaptively learns which low-rank components to remove in order to achieve a strong performance-compression trade-off.
We show that a simple magnitude based pruning strategy is sufficient to accomplish strong results. In addition, we further increase performance via an improved \lzero regularization~\citep{louizos2017learning} technique which uses an augmented Lagrangian method to directly control the final compression level of the model. Our method, which we refer to as FLOP (\textbf{F}actorized \textbf{Lo}w-rank \textbf{P}runing) applies to any matrix multiplication.

Pruning large language models introduces unique challenges with the handling of large input and output layers. Although our method is generic, it is particularly well suited to this task. In particular, we show that \ourmethod can dynamically learn the embedding dimensions of different word clusters, effectively extending the idea of adaptive embeddings and softmax~\citep{grave2017efficient,baevski2018adaptive}.
Since these embedding layers take a significant amount of parameters in the language models, learning flexible dimensions instead of specifying them manually results in a more optimal trade-off between parameter reduction and performance.

We evaluate our method on common language modeling and language understanding tasks including the Wiki-103, Enwiki8 and GLUE benchmarks, and by testing our method on both recurrent networks and Transformer~\citep{vaswani2017attention}.

Our results demonstrate that factorization based pruning significantly outperforms block-structured pruning and even surpasses unstructured pruning, while using our improved \lzero regularization further improves the performance in most cases.
When pruning a large word-level language model with adaptive embeddings for example, our method achieves 50\% compression while losing only 0.8 perplexity.
Moreover, our method is able to achieve over 2x speed-up during both training and inference with no additional hardware or software requirements.
Our method will be released as a Pytorch~\citep{paszke2017automatic} library.

%% file: text/related.tex
The development of model compression techniques can be categorized into three areas of research: 
weight pruning \citep{han2015learning,zhu2017prune}, knowledge distillation \citep{ba2014deep, hinton2015distilling, kim2016sequence}, and quantization \citep{gong2014compressing,ZhuHMD17,shen2019q}. 

Recent efforts have successfully applied compression on various architectures and NLP applications, such as pruning multi-head attentions for machine translation~\citep{voita-etal-2019-analyzing}, learning adaptive embeddings and softmax layers for language models~\citep{grave2017efficient,baevski2018adaptive,li2018slim,variani2019west}, and compressing BERT models via distillation~\citep{chia2019transformer,jiao2019tinybert,sanh2019distilbert,sun-etal-2019-patient,tsai-etal-2019-small,turc2019well}.
Only one of the compression techniques such as distillation has been used in these works for simplicity.
However, these techniques can be combined to achieve greater compression~\citep{han2015deep,shangguan2019optimizing}.
Our pruning method is compatible with quantization and distillation, as it can be applied to compress any matrix multiplication in a network.

Previous work has considered different weight pruning approaches such as
unstructured pruning based on magnitude \citep{narang2017exploring,frankle2018the}, dropout \citep{gale2019state,fan2019reducing,molchanov2017variational}, and structured pruning \cite{wen2017learning,louizos2017bayesian}.
Model weights are often removed via thresholding and $l_1$ regularization during the pruning process~\citep{narang2017block,liu2018efficient}.
Our method differs from previous work by using low-rank parameterization for compression.
Furthermore, we extend \lzero regularization using an augmented Lagrangian optimization method to control the final model size.

%% file: text/method.tex
\section{Background}
We formalize the task of model pruning as an end-to-end learning problem with \lzero regularization, following the prior work of \citet{louizos2017learning}.

Consider a given neural network model $f(\cdot; \param)$ parameterized by $\param = \{ \theta_j \}_{j=1}^{n}$,
where each $\theta_j$ represents an individual parameter weight or a block of weights (e.g. a column of a weight matrix) and $n$ denotes the number of blocks.
A pruning strategy of the model can be parameterized by introducing additional binary variables $\mask = \{ z_j \}_{j=1}^{n}$ such that $z_j \in \{0, 1\}$ and
\begin{align*}
\tilde{\param} = \param\odot \mask \quad\qquad \forall j \ \ \tilde{\theta}_j = \theta_j \, z_j .
\end{align*}
Here $\tilde{\param} = \{ \tilde{\theta}_j \}$ denotes the set of model parameters after pruning and its \lzero ~norm, $\|\tilde{\param}\|_0 = \sum_{j=1}^n z_j$, measures the effective size of the pruned model.

The choice of binary variables $\mask$ can be regulated by some prior distribution and optimized given the training data.
That is, let $q_j(z)$ be the density function of the learnable prior of $z_j$.
The optimization objective during training can be formulated as minimizing the expected training loss
\begin{align}
  \mathbb{E}_{\textbf{z}} \left[ \, \frac{1}{D} \sum_{i=1}^D \mathcal{L}\left(\mathbf{x}_i, \mathbf{y}_i; \tilde{\param} \right) + \lambda \|\tilde{\param}\|_0 \, \right], 
\end{align}
where $\{\mathbf{x}_i, \mathbf{y}_i\}_{i=1}^D$ are training examples, $\mathcal{L}$ is the training loss function and $\lambda > 0$ is a constant hyper-parameter for \lzero ~norm regularization encouraging the model to be sparse. 
Note that in practice optimizing this objective is intractable due to the discrete nature of $z_j$ and an exponential number of $2^n$ choices.

The key to the method of~\citet{louizos2017learning}, called the re-parameterization trick, enables $\mask$ to be differentiable and jointly trained with the model parameter $\param$.
Specifically, the random variables $\mask$ are relaxed as continuous variables distributed within the interval $[0, 1]$.
In addition, instead of learning the probability density function $q_j(z)$, the re-parameterization trick proposes to learn the inverse of the cumulative density function (CDF).
Note that if $G()$ is the inverse of CDF for a variable $z$, then $z$ can be easily sampled by first sampling $u \sim U(0, 1)$ and computing $z = G(u)$.
Assuming the inverse CDF function is parameterized by some learnable parameters $\bm{\alpha} = \{\alpha_j\}_{j=1}^n$ and the function $G(\cdot; \bm{\alpha})$ is differentiable, we obtain an overall end-to-end learning objective,
\begin{align}
& \min_{\param, \bm{\alpha}}\ \mathbb{E}_{\,\bm{u} \sim U(0,1)} \left[ \frac{1}{D} \sum_{i=1}^D \mathcal{L}(\mathbf{x}_i, \mathbf{y}_i; \tilde{\param}) + \lambda \|\tilde{\param}\|_0 \right], \nonumber \\
& \ z_j = G(u_j ; \alpha_j),  \ \ \forall j = 1\cdots n
\end{align}
where $\mathbf{u} = \{u_1, \cdots, u_n\}$ denotes the iid samples from the uniform distribution. 
Since $\mask$ is now the output of the parameterized function $G(\cdot; \bm{\alpha})$ and is used as an intermediate representation for the neural network (with $\tilde{\param} = \param \odot \mask$), gradient based optimization methods can perform gradient updates for $\param$ and $\bm{\alpha}$.

Following previous work, we choose the Hard Concrete distribution for the random variables $\mask=\{z_j\}$.
The inverse of CDF $G(\cdot; \bm{\alpha})$ of this distribution is defined as follows
\begin{align*}
& \mathbf{u} \sim U(0,1) \\
& \ \ \mathbf{s} = \text{sigmoid}(\log \mathbf{u} - \log(1-\mathbf{u}) + \bm{\alpha}) \\
& \ \ \bar{\mathbf{s}} = \mathbf{s} \times (r - l) + l \\
& \ \ \mask = \min(1, \max(0, \bar{\mathbf{s}}))
\end{align*}
where $l < 0$ and $r > 1$ are two constants used to `stretch` the sigmoid outputs $\mathbf{s}$ into the interval $(l, r)$, and the final outputs $\mask$ are rectified into $[0,1]$.
The stretch-and-rectify process has the effect of assigning a significant portion of probability mass on the integer values $\{0, 1\}$, which makes it a good relaxation of the binary (Bernoulli) distribution.
During training, we sample $\mathbf{u}$ and compute $\mask$ and the loss $\mathcal{L}()$ for each training batch.
The expected \lzero~norm regularization can be separately computed via a closed form
\begin{align}
\mathbb{E}\left[ \|\tilde{\param}\|_0 \right] &= \sum_{j=1}^n \mathbb{E}\left[ z_j > 0\right] \nonumber \\
&= \sum_{j=1}^n \text{sigmoid}\left(\alpha_j - \log\frac{-l}{r}\right) 
\end{align}
which is differentiable as well.

\section{Method}
In this section, we introduce \ourmethod, an improved structured pruning method.
\ourmethod proposes a different parameterization of the weight matrices using low-rank factorization. In addition, we introduce a revised optimization objective that allows for an explicit control of the compression size.

\subsection{Structured Pruning using Factorization}
In weight pruning, a key choice is how we define parameter blocks $\theta_1, \cdots, \theta_n$ to achieve the most effective pruning results.
One obvious method is to prune each individual parameter weight, which often retains strong performance but poses challenges to achieve a computation speedup given unstructured sparse matrices.

Structured pruning chooses to remove groups of consecutive parameters as a remedy.
For example, consider a fully connected layer which performs a multiplication $\mathbf{W}\mathbf{x}$ for an input feature $\mathbf{x}\in \R^d$ and weight matrix $\rmW \in \R^{d'\times d}$.
One popular method, sometimes referred to as neuron or input feature pruning, consists of adding the sparsity variables as a sparse diagonal matrix $\mathbf{G}=\text{diag}(z_1,\cdots, z_d)$ to the multiplication, i.e., $\mathbf{WGx}$.
This effectively removes the subset of the columns in $\mathbf{W}$ with $z_k=0$, where $k$ is the column index.

In practice, this method produces significant speedups at both training and inference time (by selecting a small subset of columns and performing matrix multiplications given much smaller matrices).
However, it is reported to achieve lower performance compared to unstructured pruning~\citep{yao2019balanced} due to more restrictive sparse patterns.

We propose to use low-rank factorization as a less restrictive, yet powerful representation and obtain parameter reduction by pruning rank-1 components.
That is, we reparameterize and factorize the matrix $\mathbf{W}$ into the product of two smaller matrices $\mathbf{W} = \mathbf{P} \mathbf{Q}$, where $\mathbf{P} \in \R^{d'\times r}$, $\mathbf{Q}\in \R^{r\times d}$ and $r \leq \min\{d, d'\}$ is the number of columns of $\mathbf{P}$ (equivalently the number of rows of $\mathbf{Q}$).
Let $\mathbf{p}_k$ and $\mathbf{q}_k$ be the $k$-th column of $\mathbf{P}$ and $k$-th row of $\mathbf{Q}$ respectively.
Since $\mathbf{W}$ is now the sum of $r$ rank-1 components $\mathbf{p}_k\,\mathbf{q}_k$, we can achieve structured pruning by introducing a pruning variable $z_k$ for each component 
\begin{align*}
    \mathbf{W} = \mathbf{PGQ} = \sum_{k=1}^r z_k \times (\mathbf{p}_k \times\,\mathbf{q}_k)
\end{align*}
where $\mathbf{G} = \text{diag}(z_1, \cdots, z_r)$ is again a diagonal matrix of pruning variables.
Intuitively, learning the factorization has the potential of keeping the most effective rank-1 components, and thereby better preserve the model performance.\footnote{It is also easy to see that input feature pruning is a special case of low-rank pruning: By fixing $\mathbf{P} = \mathbf{W}$ and $\mathbf{Q} = \mathbf{I}$, $\mathbf{P}\mathbf{G}\mathbf{Q} = \mathbf{W}\mathbf{G}\mathbf{I} = \mathbf{W}\mathbf{G} $.}

After training, only columns and rows corresponding to non-zero diagonal values need to be stored, resulting in much smaller (but still dense) matrices. The nonzero values of $\mathbf{G}$ can be absorbed into either $\mathbf{P}$ or $\mathbf{Q}$. 
The computation boils down to simple matrix multiplications at inference time, maximizing efficiency on common hardware. 
Unlike unstructured pruning, we need not store the indices of the sparse weights, resulting in greater memory savings.

\subsection{Pruning Adaptive Embedding and Softmax Layer}
The input embedding and softmax output layer can take the vast majority of parameters in a language model when the vocabulary size is large.

Previous work have considered various techniques that are specifically tailored to compress the embedding and softmax layer. 
For instance, the adaptive embedding and softmax methods of ~\citet{grave2017efficient,baevski2018adaptive} have been shown to achieve impressive results in preserving perplexity while significantly reducing the total number of embedding parameters.

We describe how \ourmethod fits naturally with these adaptive methods, giving them more potential.
The core idea behind the adaptive methods is to apply different embedding dimensions and projections to different word clusters.
Consider the recent method of \citet{baevski2018adaptive} without loss of generality.
Let $i \in \{1,\cdots, C\}$ denotes the indice of the i-th word cluster (sorted based on word frequency).
Two parameter matrices $\rmE_i \in \R^{n_i\times d_i}$ and $\rmO_i \in \R^{d_i \times d}$ are introduced for the i-th cluster, where $n_i$ is the number of words in the cluster, $d$ is the original embedding dimension and $d_i$ is the reduced word dimension for this cluster.
In other words, each word embedding in this cluster has dimension $d_i$ but are projected back into dimension $d$ using a projection $\rmO_i$ (and vise versa). This is in indeed a low-rank factorization
\begin{align*}
    \tilde{\rmE}_i = \rmE_i \, \rmO_i \; \in \; \R^{n_i\times d}
\end{align*}
for an underlying embedding matrix $\tilde{\rmE}_i$. 
While the reduced dimensions $\{d_i\}_{i=1}^C$ usually have to be manually specified, our method automatically learns separate diagonal pruning mask $\rmG_i$ for each cluster, i.e. $\tilde{\rmE_i} = \rmE_i \rmG_i \rmO_i$.
During training and pruning, it \emph{adaptively} learns to adjust the parameter budget of each word cluster based on what is needed to achieve good performance.
Unsurprisingly, our method prunes most of the dimensions for rare words, which is consistent with the empirical choice made in prior work.

\subsection{Augmented Lagrangian Method}
Our method can be implemented with a magnitude based pruning strategy, or directly trained with the training objective~(2) which uses an \lzero~regularization $\lambda \|\tilde{\param}\|_0$ to promote weight pruning.
One limitation of this regularization however is the lack of effective control on the size of the pruned model.
For instance, we observe that training with the same $\lambda$ could converge to very different model sizes when using slightly different learning rates or pruning schedules.
This can be problematic because a desired model size or parameter budget is often needed in many real-world applications.

We make use of an Augmented Lagrangian method to overcome this training limitation.
Lagrangian relaxation methods have been explored in many NLP problems~\citep{bastings2019interpretable,martins2011augmented,flanigan2014discriminative,rush2010dual}.
We use the following Lagrangian variant for our task -- 
Let $t$ be the target model size and $s(\bm{\alpha})$ be the expected model size determined by the Hard Concrete parameter $\bm{\alpha}$. 
Note $s(\bm{\alpha})$ can be computed based on Eq~(3) by multiplying $\mathbb{E}\left[z_j >0\right]$ with the size of the $j$-th parameter block.
Our Augmented Lagrangian method imposes an equality constraint $s(\bm{\alpha})=t$ by introducing a violation penalty,
\begin{align*}
g(\lambda, \bm{\alpha}) = \lambda_1 \cdot (s(\bm{\alpha})-t) + \lambda_2 \cdot (s(\bm{\alpha})-t)^2
\end{align*}
where $\lambda_1, \lambda_2 \in \mathbb{R}$ are two Lagrangian multipliers that will be jointly updated during training.
The overall training optimization is an adversarial game,
\begin{align*}
\max_{\lambda_1, \lambda_2}\, \min_{\param, \bm{\alpha}}\, \mathbb{E}_{\mathbf{u}} \left[\frac{1}{D}\sum_{i=1}^D\mathcal{L}(\mathbf{x}_i,\mathbf{y}_i;\tilde{\param})\right] + g(\lambda, \bm{\alpha}).
\end{align*}
The updates of $\lambda_1$ and $\lambda_2$ would always increase the training loss unless the equality constraint is met, which gives us the desired model size.

We gradually increase the target size $t$ at a linear rate during the process of pruning training. 
That is, given the desired size $t_\text{max}$, we set the sparsity at $k$-th pruning iteration as
\begin{align*}
t_k = \min(1, \frac{k}{m}) \cdot t_\text{max}
\end{align*}
where $m$ is a hyperparameter specifying the number of annealing steps. 

We perform joint gradient updates for the model parameters $\bm \theta$, $\bm \alpha$ as well as the Lagrangian multipliers $\lambda_1, \lambda_2$. For each training batch, we sample the pruning mask $\mask=\{z_1,\cdots, z_n\}$ and share it across the training examples within the batch.
Since the pruning mask is shared, we can select parameters that are only active for the current batch and compute smaller matrix multiplications in forward and backward passes.
This results in training speedup when $\mask$ becomes sparse.

\subsection{Inference}
During training, the prune mask is a random variable drawn from the Hard Concrete distribution. 
At inference time, however, we must use a deterministic, fixed mask $\mathbf{z}$ for each weight matrix to obtain the compressed factorization matrices $\mathbf{P}$ and $\mathbf{Q}$ (by keeping i-th low-rank component if $z_i >0$).
We do so by computing the expected value of each $z_i$ in $\mathbf{z}$ using Eq.(3) described in Section 3, and then keeping the top values of $\{z_1, \cdots, z_n\}$ and clipping the rest to zero, as to match the \lzero norm (i.e. the compression level).

%% file: text/results.tex
\newcommand\Tstrut{\rule{0pt}{2.6ex}}         
\newcommand\Bstrut{\rule[-0.9ex]{0pt}{0pt}}   

\section{Experimental Setup}

\paragraph{Tasks}
We evaluate the performance of our method on language modeling and BERT fine-tuning.
Specifically, we consider the following task setup.
\begin{enumerate}
    \item Recurrent word-level language models on the Wiki-103 dataset. We adopt SRU~\citep{lei2018simple} as the recurrent architecture and tied adaptive embedding and softmax layers~\citep{baevski2018adaptive}. Our base model consists of 12 recurrent layers, 100M parameters in total. About 50\% of the parameters are used for the adaptive layers. 
    \item Recurrent character-level language models on the Enwik8 dataset. We use the same SRU architecture. The base model uses 6 recurrent layers and 35M parameters in total.
    \item Transformer-XL model on the Enwik8 dataset. 
    We use the 12-layer base model from ~\citet{dai2019transformer} containing 41M parameters. We introduce pruning for the matrices in the self-attention layers as well as those in the feed-forward layers. For factorization based pruning, we choose the starting rank $r$ for each matrix such that the number of parameters remain the same as the unfactorized model\footnote{In effect, we set $r = d_1 d_2/(d_1+ d_2)$, where $d_1, d_2$ are the dimensions of the original weight matrix.}.
    \item BERT fine-tuning on several classification benchmarks benchmark~\citep{socher-etal-2013-recursive, dolan-brockett-2005-automatically, cer-etal-2017-semeval, wang2018glue}. In this experiment, we use the pre-trained RoBERTa base model by \citet{liu2019roberta}.

\end{enumerate}
We extend the implementation of Transformer, SRU and the adaptive embedding / softmax layers to support factorization based pruning (and other baselines).

\paragraph{Baselines}
We compare with the following unstructured, structured and/or factorization based pruning baselines.
\begin{itemize}

\item {\bf FAC} which trains low-rank factorized models from scratch by reducing all dimensions with the same ratio to get the desired compression.
\item {\bf NP-\lzero} \citep{louizos2017learning} which adopts \lzero regularization and performs neuron pruning (i.e. removing input features and columns of weight matrices). No factorization is used for this baseline. We add the Augmented Lagrangian optimization similar to \ourmethod to achieve the exact desired compression.
\item {\bf AGP} \citep{zhu2017prune} which gradually prunes individual parameters based on the weight magnitude. AGP is one of the state-of-the-art unstructured pruning methods.
We use the implementation provided in the Nervana Distiller library~\citep{neta_zmora_2018_1297430}.
\item {\bf FLOP-AGP} is a variant of our full method that prunes low-rank components, but uses magnitude-based gradual pruning on the diagonal mask $\rmG$ instead. We also tune \lone regularization on the masks to encourage sparsity, similar to~\citet{narang2017block}.
\end{itemize}
These baselines serve as competitive pruning alternatives, and also provide data points for us to isolate the effectiveness of sub-components of our method, such as low-rank factorization and \lzero pruning.
All methods use the same training configurations such as learning rate and dropout.
We tune hyper-parameters related to pruning such as compression scheduling and the learning rate of Lagrangian variables for each method.
More training and implementation details are provided in the appendix.

\section{Results}

\paragraph{Word-level Language Model} 
Table~\ref{tab:wt103} presents the results of \ourmethod as well as the baseline methods.
The SRU base model (unpruned) achieves a test perplexity of 24.5, being a strong starting point and competitive with top-performing models such as Transformer~\citep{dai2019transformer}.

\begin{table}[!t!]
    \centering
    \begin{tabular}{lrcc}
    \hline
    \bf Method & \bf Size & \bf Compress & \bf PPL \\
    \hline
    Trans. (\citeauthor{dai2019transformer}) & 151M & - & 24.1\\
    SRU (base) & 100M & - & 24.5 \\
   \hdashline
    FAC & 50M & 50\% & 28.2 \\
    AGP & 50M & 50\% & 25.7 \\
    NP-\lzero & 51M & 50\% & 26.7 \\
    \ourmethod-AGP & 51M  & 50\% & 25.6\\
    \ourmethod-\lzero & 50M & 50\% & \bf 25.3 \\
   \hdashline
    FAC & 30M & 70\% & 31.0 \\
    AGP & 30M & 70\% & 28.4 \\
    NP-\lzero & 31M & 70\% & 31.3 \\
    \ourmethod-AGP & 31M & 70\% & 28.1 \\
    \ourmethod-\lzero & 30M & 70\% & \bf 27.7 \\
    \hdashline
    FAC & 21M & 80\% & 35.2 \\
    AGP & 20M & 80\% & 32.6 \\
    NP-\lzero & 18M & 80\% & 39.1\\
    \ourmethod-AGP & 21M & 80\% & \bf 31.3 \\
    \ourmethod-\lzero & 21M & 80\% & 31.9 \\
    \hline
    \end{tabular}
    \caption{Comparison of \ourmethod and all baselines on the Wiki-103 dataset. We report test perplexity (PPL) at three different compression levels. All methods use adaptive embedding and softmax layers.}
    \label{tab:wt103}
\end{table}

The pruning results conform to our expectations that pruning a large model is consistently better than training a small model from scratch, and using low-rank based pruning yields better performance than removing matrix columns and input features. 
\ourmethod exceeds the performance of FAC, NP-\lzero and AGP baselines at all compression levels tested.
The performance of FLOP-AGP, especially in comparison with its unstructured counterpart AGP, highlights the effectiveness of factorization based pruning. Moreover, we achieve a test perplexity (PPL) of 25.3 with FLOP-\lzero method, a loss of 0.8 perplexity score, while removing 50\% of the model parameters. 
This result is impressive since our base model adopts the adaptive word embedding and softmax layers, which already reduce the model size significantly.

\begin{figure}[!t!]
    \includegraphics[width=3.1in]{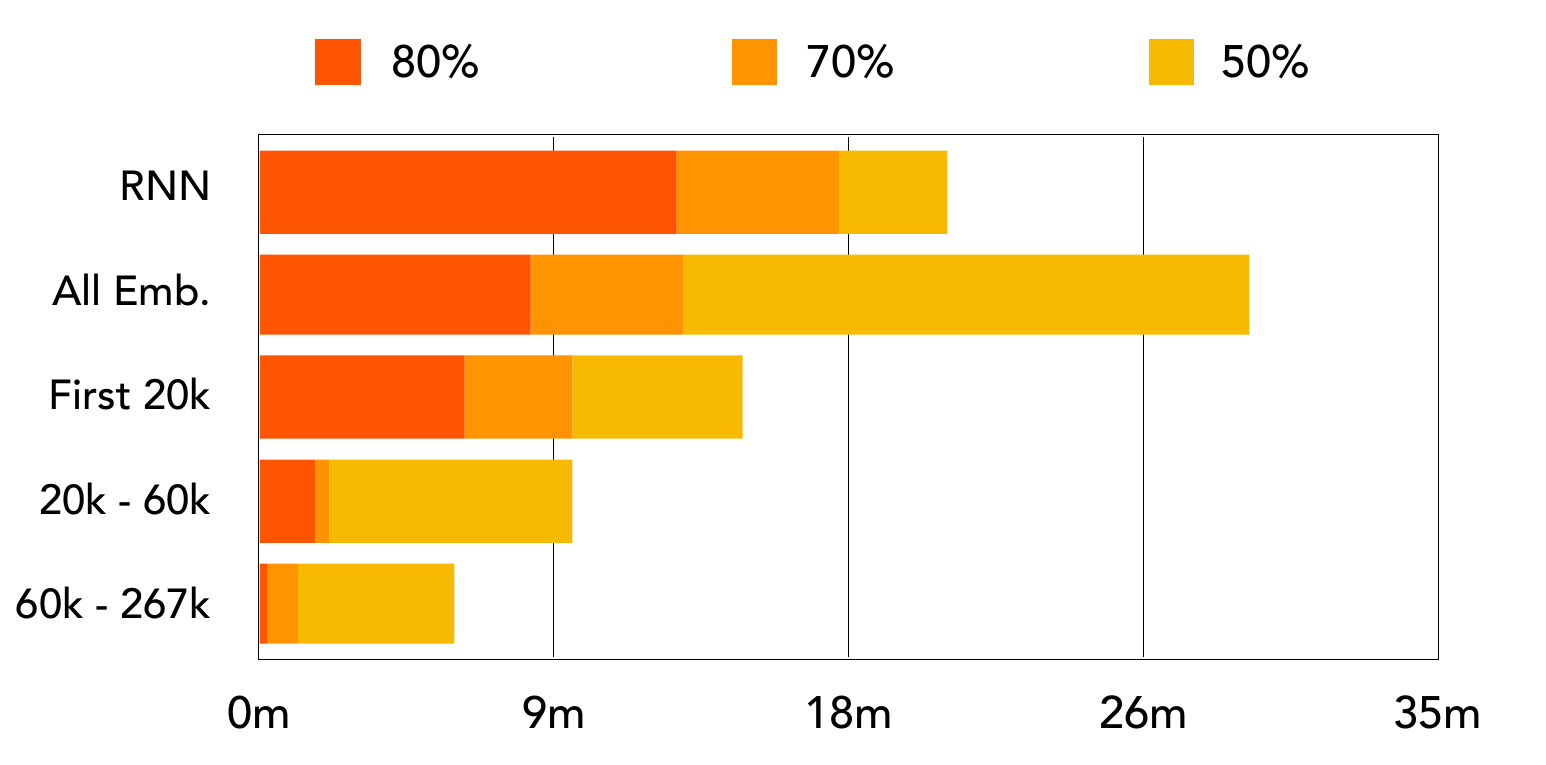}
    \caption{Number of parameters used in the RNN and adaptive embeddings at different compression levels. We also show the number of parameters used for the most, second most and least frequent words.}
    \label{fig:prune_breakdown}
\end{figure}

Figure~\ref{fig:prune_breakdown} illustrates how our method adaptively controls the size of different model components.
We show the overall size of recurrent encoder and adaptive embedding layers at the compression levels tested, and break down the use of parameters within three word clusters based on their frequency.
\ourmethod learns to prune the dimension more aggressively for less-frequent words.

This result showcases the benefit of adaptively reducing word dimensions.

\begin{table*}[!h!]
\vspace{0.05in}
\begin{center}
\begin{tabular}{ccccccc}
\hline
\bf Parameters & \bf Compression &\multicolumn{1}{@{~~~}c@{~~~}}{\bf SST2}  &\multicolumn{1}{@{~~~}c@{~~~}}{\bf MRPC} &\multicolumn{1}{@{~~~}c@{~~~}}{\bf STS-B} &\multicolumn{1}{@{~~~}c@{~~~}}{\bf QNLI} &\multicolumn{1}{@{~~~}c@{~~~}}{\bf Average}\\
\hline
125M & 0\% & 92.43 &	90.9 &	90.22 &	89.77 & \textbf{90.83} \\
80M & 35\% & 92.09 & 88.61 & 88.18 & 89.05 & \textbf{89.48}
\\ \hline
\end{tabular}
\end{center}
\vspace{-0.05in}
\caption{Compression on downstream fine-tuning}
\label{table:finetune}
\end{table*}

\begin{table}[!t!]
    \centering
    \begin{tabular}{lrcc}
    \hline
    \bf Method & \bf Size & \bf Comp. & \bf BPC \\
    \hline
    LSTM (\citeauthor{wu2016milstm}) & 17M & - & 1.44 \\
    QRNN (\citeauthor{merity2018analysis}) & 26M & - & 1.33 \\
    SRU (base) & 35M & - & 1.24 \\
   \hdashline
    FAC & 11M & 70\% & 1.33 \\
    AGP & 11M & 70\% & 1.27\\
    NP-\lzero & 11M & 70\% & 1.31 \\
    \ourmethod-AGP & 11M & 70\% & 1.27\\
    \ourmethod-\lzero & 11M & 70\% & \bf 1.25 \\
   \hdashline
    FAC & 8M & 80\% & 1.38 \\
    AGP & 8M & 80\% &  1.29 \\
    NP-\lzero & 8M & 80\% & 1.34 \\
    \ourmethod-AGP & 8M & 80\% & 1.29 \\
    \ourmethod-\lzero & 8M & 80\% & \bf 1.27 \\
    \hdashline
    FAC & 4M & 90\% & 1.47 \\
    AGP & 4M & 90\% & 1.35 \\
    NP-\lzero & 4M & 90\% & 1.43\\
    \ourmethod-AGP &  4M & 90\% & 1.34 \\
    \ourmethod-\lzero & 4M & 90\% & \bf 1.33 \\
    \hline
    \end{tabular}
    \caption{Comparison of \ourmethod and all baselines on the Enwiki8 dataset. We report bits-per-character (BPC) on the test set. We also include previous reported results of recurrent language models on this dataset as additional data points.}
    \label{tab:enwik8}
\end{table}

\begin{table}[!t!]
    \centering
    \begin{tabular}{lrcc}
    \hline
    \bf Method & \bf Size & \bf Compress & \bf BPC \\
    \hline
    Trans-XL (base) & 41M & - & 1.08 \\
   \hdashline
    FAC & 8M & 80\% & 1.20 \\
    AGP & 8M & 80\% & 1.14 \\
    \ourmethod-AGP & 8M & 80\% & 1.17\\
    \ourmethod-\lzero & 8M & 80\% & \bf 1.13 \\
       \hdashline
    \ourmethod-AGP & 4M & 90\% & 1.25 \\
    \ourmethod-\lzero & 4M & 90\% & \bf 1.17\\
    \hline
    \end{tabular}
    \caption{Results of pruning Transformer-XL models on the Enwiki8 dataset. We report bits-per-character (BPC) on the test set.}
    \label{tab:enwik8tf}
\end{table}

\paragraph{Char-level Language Model}
Table~\ref{tab:enwik8} shows the results of pruning character-level language models. 
Our base model achieves a test bits-per-character score (BPC) of 1.24, which is comparable with previous reported results of RNN-based models.

As shown in Table~\ref{tab:enwik8}, we again see the benefit of low-rank pruning, matching or improving on unstructured pruning. Furthermore, \ourmethod-\lzero obtains the best performance across all pruning levels. Notably, we achieve a perplexity of 1.25 at 70\% compression, nearly matching the un-compressed model at 1.24. 

Table~\ref{tab:enwik8tf} presents the results of pruning 12-layer Transformer-XL models on the Enwik8 dataset.
We compare FAC, unstructured AGP, FLOP-AGP and FLOP-\lzero at 80\% compression level, and also report the result of FLOP variants at 90\% compression.

FLOP-\lzero outperforms other methods in comparison.
In addition, it is able to achieve 1.17 BPC using 4M parameters, showcasing the effectiveness of our method when applied to another neural architecture.

\paragraph{BERT on Classification Tasks}
Finally, we demonstrate that our method can also be applied to language model fine-tuning on downstream tasks. 
We use the RoBERTa base model in this experiment.
Since the model was pretrained without matrix factorization, we first compute the singular value decomposition of each matrix and then introduce the pruning mask in between the resulting factored matrices. 
Note that this procedure temporarily increases the total number of parameters. 
We compare here the final number of parameters to the initial number pre-factorization.

Our results are shown in in Table~\ref{table:finetune}. We are able to conserve nearly 99\% of the performance while reducing the number of parameters by 35\%. Our target compression level is limited by the fact that the embedding layers consist of a significant portion of the remaining parameters. 
As demonstrated in the previous experiment on Wiki-103, we believe that higher levels of compression could be obtained by factorizing the embedding layer, similar to \citet{lan2019albert}.

%% file: text/analysis.tex
\label{sec:analysis}
In this section, we perform an analysis of several aspects of our method.


\begin{table*}[!t]
\centering
\begin{tabular}{c@{~~~~~}c@{~~~~~}ccccc}
\hline
\bf Variants  & \bf Size & ~~~~~0\%~~~~~ & 70\% & 80\% & 85\% & 90\% \\
 \hline
 \multirow{2}{*}{NP-\lzero} & \multirow{1}{*}{37M} & 1.30  & 1.31 (-0.8\%) & 1.34 (-3.2\%) & 1.37 (-5.4\%) & 1.43 (-10.0\%) \\
 & \multirow{1}{*}{66M} & 1.25 & 1.28 (-2.4\%) & 1.31 (-4.8\%) & 1.32 (-5.6\%) & 1.37$\ \ $ (-9.6\%) \\
 \hline
 \ourmethod-\lzero & \multirow{1}{*}{35M} & {\bf 1.24} & {\bf 1.25 (-0.8\%)} & {\bf 1.27 (-2.4\%)} & {\bf 1.29 (-4.0\%)} & {\bf 1.33$\ \ $ (-7.3\%)} \\
 \hline
\end{tabular}
\caption{Further comparison between factorization-based pruning \ourmethod and input feature pruning NP-\lzero~\citep{louizos2017learning} using 6-layer SRU models and the Enwiki8 dataset. We show BPC at different compression levels and the loss of performance relative to the un-compressed model. Factorization results in less decrease in relative and absolute performance.}
\label{table:factorization_analysis}
\end{table*}

\begin{figure*}[!t!]
    \centering
    \begin{tabular}{ccc}
    \includegraphics[height=1.45in]{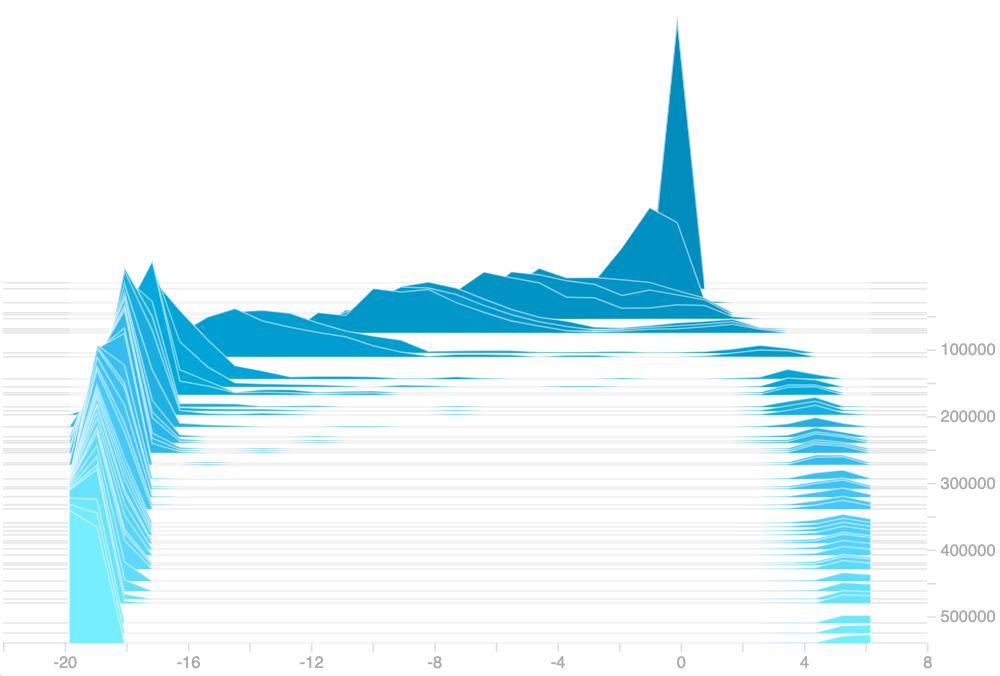}
    &
    ~~~~
    &
    \includegraphics[height=1.45in]{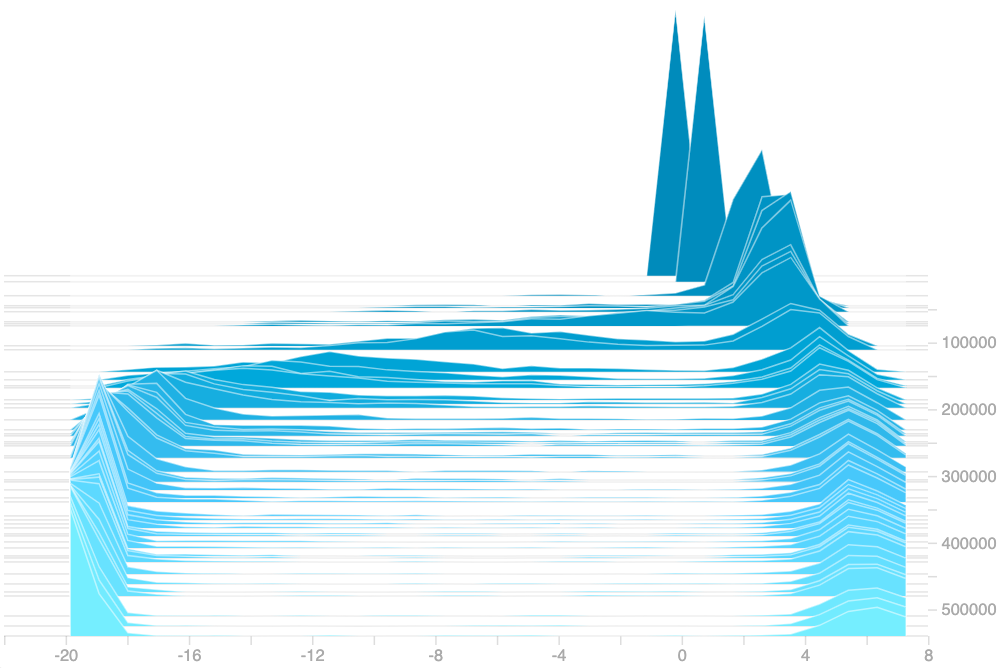}
    \end{tabular}
    \caption{Histograms of HardConcrete parameters during training. We show the changes of histograms for the first SRU layer (left figure) and the last layer (right figure). We compute the histogram every 3,000 training steps.}
    \label{fig:hardconcrete_dynamics}
\end{figure*}

\paragraph{Factorization}


One of the key hypotheses outlined in this paper is that pruning input dimensions (equivalently rows or columns of weight matrices) is a more restrictive form of pruning compared to our factorization based strategy.
However, one could also argue that the factorization method works better simply because the hidden size can be initially set much larger than an unfactorized model, not because of pruning itself.
For instance, the SRU model used by the unfactorized NP-\lzero baseline has hidden size 1536, while with factorization other baselines with a similar parameter budget use a hidden size of 3056.
To avoid potential unfair comparison, we also train a large model with hidden size $2048$ containing 90\% more parameters, and apply the NP-\lzero baseline.
This larger model obtains 1.25 BPC which is on par with the factorized base model used in previous experiments.

Table~\ref{table:factorization_analysis} compares the pruning performance of \ourmethod and NP-\lzero at four compression levels.
We show the test BPC and the loss of performance relative to the model without pruning.
These results further substantiate our hypothesis -- 
factorization based pruning is able to retain relative model performance much more effectively than input feature pruning.

\paragraph{Speed analysis}
Thanks to its structured nature, \ourmethod can achieve significant computation speedup.
As shown in Table~\ref{table:sru_inference}, we achieve an inference speedup ranging from 1.5x to 2.2x for the compression levels tested, using CPUs.
Similar speedups of up to 2.4x are also observed using GPUs during training.
On the contrary, the computations of unstructured sparse matrices are harder to optimize. For models obtained using unstructured AGP, we experimented with the sparse matrix multiplication routine provided in Pytorch \citep{paszke2017automatic} and a recent linear algebra compiler~\citep{kjolstad2017tensor}, but were unable to achieve a speedup.

\begin{table}[!t]
\label{sru-time}
\begin{center}
\begin{tabular}{cccc}
\hline
\bf Size & \bf Compress  &\multicolumn{1}{c}{\bf Time (s)} &\multicolumn{1}{c}{\bf Speedup} \\
\hline
\Tstrut
35M & 0\% & 0.39 & 1.0x \\
8M & 80\% & 0.21 & 1.9x\\
4M & 90\% & 0.18 & 2.2x \\
\hline
41M & 0\% & 1.33 & 1.0x \\
8M & 80\% & 0.87 & 1.5x \\
4M & 90\% & 0.82 & 1.6x
\\ \hline 
\end{tabular}
\end{center}
\caption{Inference timing measurements of character-level language model using SRU (top block) and Transformer-XL (bottom block).}
\label{table:sru_inference}
\end{table}

\paragraph{Learning dynamics}
Figure~\ref{fig:hardconcrete_dynamics} demonstrates the training dynamics of the HardConcrete distribution.
We plot the histogram of HardConcrete parameters $\bm{\alpha}$ after every few thousands of training iterations.
A negative value of $\bm{\alpha}$ indicates that the associated parameter is likely to be pruned while a positive value indicates the opposite.
The magnitude of the value reflects the certainty of the pruning decision.
As illustrated by the figure, the distribution of $\bm{\alpha}$ becomes bi-modal after initial exploration.
Certain parameters within each layer are completely pruned while others are kept with (almost) absolute certainty.
In addition, the dynamics vary across different layers.
For instance, for SRU the first recurrent layer gets pruned more aggressively than the last layer.

%% file: text/conclusion.tex
In this work, we present a generic structured pruning method based on adaptive low-rank factorization. We systematically evaluate the performance of this method on large language models. We show that our method can provide significant speedups and compression rates on large models while losing minimal performance compared to other methods, including unstructured magnitude pruning. This work contributes to reducing the growing overhead of large language models, and shines a light on the role of model capacity in language modeling.

%% file: text/acknowledgement.tex
We would like to thank ASAPP Inc. for making this work possible. We would also like to thank Hugh Perkins, Sam Bowman, Nicholas Matthews, Josh Shapiro and the other members of the Language Technology and Research teams who helped review this work and contributed their thoughts throughout the project. We would also like to thank the EMNLP reviewers and area chair for their helpful comments.

%% file: text/appendix.tex
\clearpage

\section{Appendix}

\subsection{Optimization details}
In our implementation, \ourmethod trains the factorized model for a number of warmup epochs and then starts pruning.
Other pruning baselines use the same warmup training process, except that FAC baseline directly trains smaller factorized model from scratch.
Recall our augmented Lagrangian training objective during pruning is, 
\begin{align*}
\max_{\lambda_1, \lambda_2}\, \min_{\param, \bm{\alpha}}\, \mathbb{E}_{\mathbf{u}} \left[\frac{1}{D}\sum_{i=1}^D\mathcal{L}(\mathbf{x}_i,\mathbf{y}_i;\tilde{\param})\right] + g(\lambda, \bm{\alpha}),  \\
g(\lambda, \bm{\alpha}) = \lambda_1 \cdot (s(\bm{\alpha})-t) + \lambda_2 \cdot (s(\bm{\alpha})-t)^2 .
\end{align*}
We gradually increase the target size $t$ at a linear rate. 
That is, given the desired size $t_\text{max}$, we set the sparsity at $k$-th pruning iteration as
\begin{align*}
t_k = \min(1, \frac{k}{m}) \cdot t_\text{max}
\end{align*}
where $m$ is a hyperparameter specifying the number of annealing steps. 

The Lagrangian multipliers are initialized to zero at the start of training.
We perform joint gradient updates for the parameters and Lagrangian multipliers at every iteration, but use and tune a different learning rate for Lagrangian multipliers.
For each training batch, we sample the pruning mask $\mask=\{z_1,\cdots, z_n\}$ and share it across the training examples within the batch.
Since the pruning mask is shared, we can select parameters that are only active for the current batch and compute smaller matrix multiplications in forward and backward passes.
This can result in training speedup when $\mask$ becomes sparse.

\subsection{Experimental Details}
Our experiments are performed using the standard train/dev/test splits of Wiki-103, Enwik8 and GLUE benchmarks.
We describe training configurations in the following paragraphs. 
Detailed experimental setup can be found at \url{https://github.com/asappresearch/flop}.

\paragraph{SRU}
Following the practice of~\citet{lei2018simple}, for the Enwik8 dataset we train a 6-layer SRU model using a batch size of 64 and an unroll length of 256.
We use a hidden size of 3056 and set the initial factorization dimension $r$ of the parameter matrices to 512.
That is, we replace each weight matrix $\mathbf{W}$ in SRU using an explicit factorization $\mathbf{PQ}$ with an inner dimension of 512.
We train the model without pruning for 30 warmup epochs, and start pruning for a maximum of 100 epochs.

For the Wiki-103 dataset, our 12-layer SRU base model uses a hidden dimension of 2048 and a factorization dimension of 512 for weight matrices in SRU.
Following~\citet{baevski2018adaptive}, the adaptive embedding layer uses 1024, 256 and 64 dimensions
respectively for the 20K most frequent words, 40K less frequent words and the rest least frequent words.
We train 50 warm-up epochs and start the pruning process for an addition of 100 epochs.
We use a batch size of 64 or 96 and an unroll length of 256.

For all SRU runs, we use inverse-square-root learning rate scheduling~\citep{vaswani2017attention} and a learning rate of $\frac{l_0}{\sqrt{d}}$ where $d$ is the hidden size and $l_0$ is the initial factor.
We set $l_0 \in \{2, 3\}$ for model parameters.
For AGP methods, we tune the start and end epoch of the compression scheduler.
For \lzero regularization, we tune the learning rate $l_0\in \{3, \cdots, 6\}$ for Lagrangian multipliers.

\paragraph{Transformer-XL}
Following \citet{dai2019transformer}, we train Transformer-XL base model using cosine learning rate scheduling.
For the 12-layer base model, we train a maximum of 200k iterations, a batch size of 48 and an initial learning rate of 0.0003 and use 8 GPUs in parallel.
For pruning runs, we train up to 300k iterations using a learning rate of 0.00025, a batch size of 32 and 4 GPUs in parallel for each run.
We use the same inverse-square-root learning rate scheduling for Lagrangian multipliers and set $l_0 \in \{0.5, 1.0, 1.5\} $.